\documentclass[letterpaper]{article} 
\usepackage{aaai2026}  
\usepackage{times}  
\usepackage{helvet}  
\usepackage{courier}  
\usepackage[hyphens]{url}  
\usepackage{graphicx} 
\usepackage{natbib}  
\usepackage{caption} 
\usepackage{algpseudocode}
\usepackage{fontawesome} 
\usepackage{algorithm}
\usepackage{amsmath}
\usepackage{booktabs}
\usepackage{multirow}
\usepackage{fontawesome}
\usepackage{marvosym}

\makeatletter
\def\thanks#1{\protected@xdef\@thanks{\@thanks
        \protect\footnotetext{#1}}}
\makeatother

\usepackage{newfloat}
\usepackage{listings}
\DeclareCaptionStyle{ruled}{labelfont=normalfont,labelsep=colon,strut=off} 
\floatstyle{ruled}
\newfloat{listing}{tb}{lst}{}
\floatname{listing}{Listing}
\title{ELSPR: Evaluator LLM Training Data Self-Purification on Non-Transitive Preferences via Tournament Graph Reconstruction}

\author{
    Yan Yu\textsuperscript{\rm 1},
    Yilun Liu\textsuperscript{\rm 2 \scalebox{1.35}{\Letter}}\thanks{\Letter ~ Corresponding author.},
    Minggui He\textsuperscript{\rm 2},
    Shimin Tao\textsuperscript{\rm 2},
    Weibin Meng\textsuperscript{\rm 2},
    Xinhua Yang\textsuperscript{\rm 2},
    Li Zhang\textsuperscript{\rm 2},
    Hongxia Ma\textsuperscript{\rm 2},
    Dengye Li\textsuperscript{\rm 3},
    Daimeng Wei\textsuperscript{\rm 2},
    Boxing Chen\textsuperscript{\rm 4},
    Fuliang Li\textsuperscript{\rm 1 \scalebox{1.35}{\Letter}}
}
\affiliations{
    \textsuperscript{\rm 1}Northeastern University, Shenyang, China\\
    \textsuperscript{\rm 2}Huawei, Beijing, China\\
    \textsuperscript{\rm 3}Tongji University, Shanghai, China\\
    \textsuperscript{\rm 4}Huawei Canada, Montreal, Canada\\
    \texttt{\ \Letter \ liuyilun3@huawei.com, lifuliang@cse.neu.edu.cn}
}

\begin{document}

\maketitle

\begin{abstract}
Pairwise evaluation of large language models (LLMs) has become the dominant paradigm for benchmarking open-ended tasks, yet non-transitive preferences—where evaluators prefer A over B, B over C, but C over A—fundamentally undermine ranking reliability. We show that this critical issue stems largely from low-quality data that contains inherently ambiguous preference pairs. To address this challenge, we propose ELSPR, a principled graph-theoretic framework that models pairwise preferences as tournament graphs and systematically identifies problematic training data. ELSPR quantifies non-transitivity through strongly connected components (SCCs) analysis and measures overall preference clarity using a novel normalized directed graph structural entropy metric. Our filtering methodology selectively removes preference data that induce non-transitivity while preserving transitive preferences. Extensive experiments on the AlpacaEval benchmark demonstrate that models fine-tuned on ELSPR-filtered data achieve substantial improvements: a 13.8\% reduction in non-transitivity, a 0.088 decrease in structural entropy, and significantly enhanced discriminative power in real-world evaluation systems. Human validation confirms that discarded data exhibit dramatically lower inter-annotator agreement (34.4\% vs. 52.6\%) and model-human consistency (51.2\% vs. 80.6\%) compared to cleaned data. These findings establish ELSPR as an effective data self-purification approach for developing more robust, consistent, and human-aligned LLM evaluation systems.
\end{abstract}
\begin{links}
    \link{Code \& Datasets}{https://github.com/yy0525/ELSPR}
\end{links}
\section{Introduction}

With the rapid advancement of large language model (LLM;~\citealt{openai2024gpt4technicalreport,qwen25max,llama31}) technology, an increasing number of models have become available, making it essential to evaluate their capabilities for selecting the most suitable one. However, existing benchmarks such as MMLU~\cite{hendrycks2020measuring} and HELM~\cite{liang2022holistic} have been shown to be insufficient for capturing performance differences in open-ended tasks~\cite{zheng2024judging_mt}.

Given the lack of definitive answers in open-ended tasks, human expert evaluation is considered the gold standard. Yet, due to its high cost and limited scalability, the mainstream approach has shifted toward using LLM-as-a-Judge for efficient evaluation. Recent studies have demonstrated that powerful LLMs, such as GPT-4, can achieve high consistency with human judgments~\cite{zheng2024judging_mt,NEURIPS2023_5fc47800}. Among various approaches, pairwise comparison has emerged as the dominant paradigm, due to its strong alignment with human preferences~\cite{NEURIPS2024_8147a43d,chen-etal-2024-humans,alpaca_eval,chiang2024chatbot,liu2024aligning,liusie-etal-2024-llm}.

\begin{figure}[htbp]
  \centering
\includegraphics[width=1\linewidth]{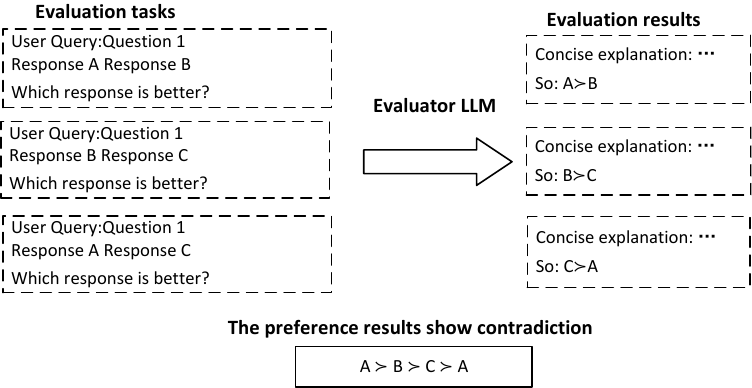}

\caption{Non-Transitive Preferences in LLM-as-a-Judge for Pairwise Comparisons (e.g., $A \succ B$, $B \succ C$, $C \succ A$).}
  \label{fig:non-transitivity} 
\end{figure}

Despite these promising results, LLM-as-a-Judge still suffers from various biases—such as position, verbosity, conformity, selection, and self-reinforcement bias—which compromise evaluation reliability~\cite{xu-etal-2024-pride,zheng2024judging_mt,koo-etal-2024-benchmarking,ye2025justice,wei-etal-2024-unveiling,choi-etal-2025-mitigating}. A particularly underexplored yet critical issue is the non-transitivity of preferences generated by evaluator LLMs (e.g., $A \succ B$, $B \succ C$, $C \succ A$), where $\succ$ denotes ``is preferred to''; that is, $A \succ B$ means $A$ is preferred over $B$. An illustration is provided in Figure~\ref{fig:non-transitivity}.
Recent research~\cite{xu2025investigating} analyzed inconsistencies in the GPT-4 evaluation outcomes when conducting pairwise comparisons between different baseline models. Building on the widely adopted AlpacaEval framework, their findings indicate that preference non-transitivity significantly undermines the robustness of evaluation systems, leading to inconsistent model rankings under various baselines and thereby compromising the reliability of LLM-based evaluators. However, their analysis is limited to observing preference non-transitivity within triplets of samples and does not extend to larger sample sets. Moreover, they do not propose an effective method to substantially reduce the degree of non-transitivity exhibited by evaluator LLMs. These limitations highlight the need for more comprehensive investigations, particularly with larger sample sets, to further explore preference non-transitivity within LLM-based evaluation frameworks.

Worryingly, this issue is also inherited by specialized evaluators such as JudgeLM, PandaLM, and Auto-J~\cite{zhu2025judgelm,wang2024pandalm,li2024generative}, which are trained through the distillation of knowledge from advanced models. The distillation process may inadvertently propagate non-transitive judgment patterns to downstream evaluators.

We hypothesize that the presence of low-quality training data may impair the transitivity of the preferences generated by the evaluator LLM. Many pairwise comparisons, particularly those from open-ended tasks, lack definitive ground truth due to their inherent subjectivity. Human annotators frequently demonstrate significant disagreement, with empirical studies reporting inter-annotator agreement rates as low as 65.7\%~\cite{alpaca_eval}, indicating that such tasks fundamentally lack universally accepted preference orderings. Consequently, training models with ambiguous and low-quality training data may introduce or exacerbate non-transitive preference relationships, undermining the development of stable and reliable evaluator LLMs. This phenomenon underscores the critical necessity of implementing robust filtering mechanisms to eliminate unreliable data points prior to model training.

In this paper, we present a novel graph-theoretic approach to assess and mitigate preference non-transitivity in evaluator LLMs. Our method, ELSPR, formulates multi-response pairwise comparisons as tournament graphs and systematically filters preference data that induce to overall preference non-transitivity. To quantify preference clarity, we introduce a new metric based on two-dimensional structural entropy of directed graphs. Experimental results demonstrate that models fine-tuned on cleaned data reduce preference non-transitivity by 13.78\% and structural entropy by 0.0879, while achieving more robust rankings in real-world evaluation systems, as evidenced by increased standard deviations in MT-bench evaluation framework metrics. Human evaluation studies further validate our approach, confirming that data inducing to overall preference non-transitivity correlates with low inter-annotator agreement among human evaluators and poor alignment between LLM evaluations and human majority votes. These findings underscore ELSPR's effectiveness in developing more reliable evaluation systems for LLMs. Specifically, our contributions are:
\begin{itemize}
    \item We introduce a graph-theoretic approach to systematically study the non-transitivity problem in preference data generated by evaluator LLM, reveal significant non-transitivity, and propose a two-dimensional structural entropy to quantify preference clarity.
    \item We propose a robust filtering methodology leveraging tournament graph theory to systematically identify and eliminate preference data that induce non-transitivity in evaluator LLMs. Our experimental results demonstrate that this approach significantly reduces preference non-transitivity, decreases structural entropy, and substantially enhances the evaluation robustness of the resulting models when deployed in real-world evaluation systems. 
    \item We empirically demonstrate through human evaluation that data causing preference non-transitivity is inherently ambiguous and low-quality, with significantly lower inter-annotator agreement (34.4\% vs. 52.6\%) and lower consistency between evaluator LLM assessments and human majority votes (51.2\% vs. 80.6\%).
\end{itemize}
\section{Related Work}
\subsection{LLM-as-a-Judge and Its Non-Transitive Preferences}
The prevailing LLM-as-a-Judge paradigm operates primarily through pairwise comparisons, as evidenced in established frameworks such as VicunaEval, AlpacaEval, and Arena-Hard~\cite{vicuna2023,alpaca_eval,DBLP:journals/corr/abs-2406-11939}. These systems collect responses generated by various LLMs for a given set of questions and then employ advanced LLMs as judges to determine preference orders between response pairs, thus evaluating the relative performance of different models. Recent studies demonstrate that even advanced models such as GPT-4 exhibit significant preference non-transitivity when used in evaluation systems, substantially undermining the reliability of evaluation outcomes. Despite systematic exploration of six distinct prompt templates, improvements in mitigating preference non-transitivity remain limited~\cite{xu2025investigating}. This issue extends to Reinforcement Learning from Human Feedback (RLHF), where non-transitive preferences significantly impair performance. Recent research suggests that this non-transitivity can hinder the learning algorithm from stably converging to the global optimum, leading to preference cycles and thus hindering the effective utilization of preference data~\cite{zhou2025extragradient}.

To address this critical challenge, we systematically investigate the underlying conditions that generate non-transitivity in LLM-as-a-Judge systems and introduce a novel methodology to mitigate such inconsistencies.

\begin{figure*}[htbp]
  \centering
  \includegraphics[width=1\textwidth]{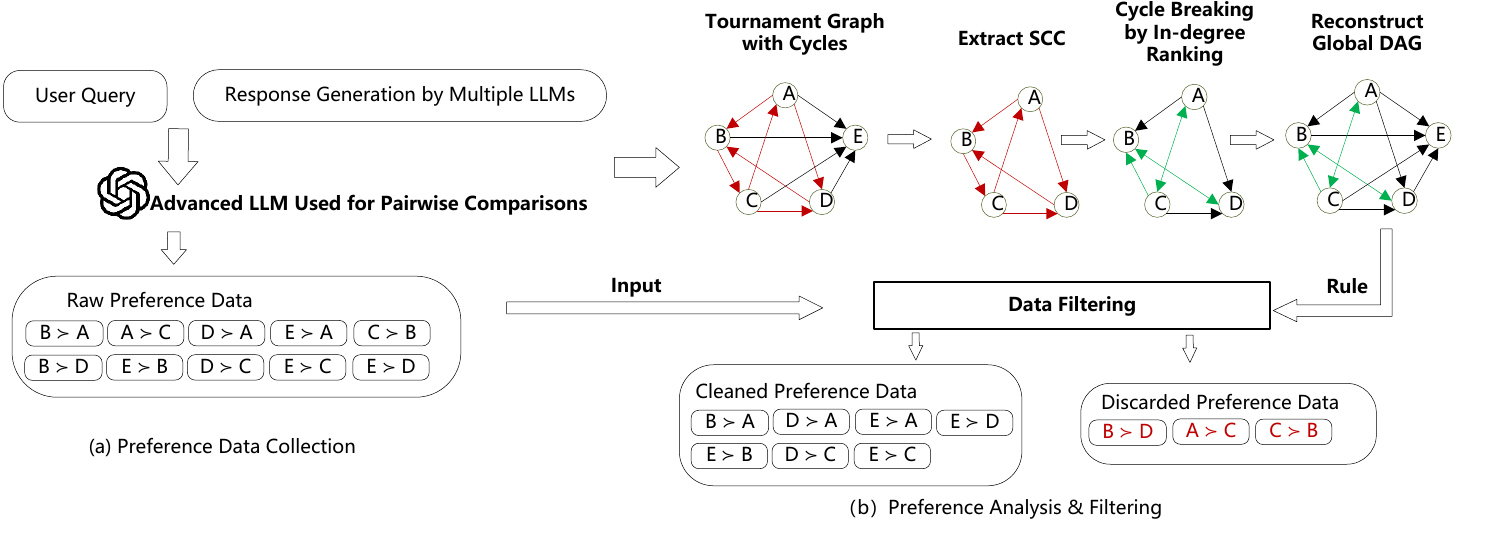}
\caption{\label{fig:overview}ELSPR (\textbf{E}valuator \textbf{L}LM training data \textbf{S}elf-purification non-transitive \textbf{P}references via tournament graph \textbf{R}econstruction) framework overview. \textbf{(a)} Raw preference data is collected via pairwise comparisons conducted by an advanced LLM. \textbf{(b)} The core analysis and filtering process: The raw data is first modeled as a tournament graph to identify cycles within SCCs. These cycles are then broken by reconstructing each SCC into a DAG based on in-degree ranking. The final global DAG serves as a rule to filter the initial raw data, separating it into a cleaned, transitively consistent training set and a discarded set of non-transitive preferences.}
\end{figure*}
\subsection{Data Selection for LLM Fine-tuning}
Previous research~\cite{chen2024alpagasus, ge-etal-2024-clustering, li-etal-2024-one} has established that extracting high-quality subsets from synthetic training sets is crucial for effective fine-tuning. This approach not only enhances the performance of the model, but also substantially reduces the computational requirements. However, this strategy remains underexplored in the domain of LLM-as-a-Judge. Existing methodologies predominantly focus on scaling model architectures or refining engineering techniques, while neglecting the potential benefits of optimizing training data quality.

\section{Methodology}
This section presents a graph-theoretic analysis method using tournament graphs to analyze pairwise preferences from evaluator LLMs. The method introduces quality analysis criteria for training data, including non-transitivity detection via strongly connected components (SCCs) and preference clarity analysis via graph entropy. Additionally, a data filtering method is proposed to mitigate non-transitive preferences. Figure~\ref{fig:overview} outlines the overall framework, with implementation details provided in the following subsections.

\subsection{Background}
\label{subsec:graph_paradigm}
\textbf{Modeling preferences generated by LLM as a tournament graph.}
For each question $q_i \in Q$, given the response set $A_i = \{a_1, a_2, ..., a_n\}$ from $n$ LLMs, we construct a tournament graph $G_i = (V_i, E_i)$ through the following procedure:
\textbf{vertices}: set $V_i = \{v_1, v_2, ..., v_n\}$ corresponds to responses $A_i = \{a_1, a_2, ..., a_n\}$. \textbf{Edges}: Defined by preferences generated by evaluator LLM:
\begin{equation}
    \begin{aligned}
        E_i = \bigcup_{\substack{1 \leq j,k \leq n \\ j \neq k}} 
        \begin{cases} 
            v_k \rightarrow v_j, & 
            \begin{aligned}[t]
            \text{if } \mathcal{J}(a_j, a_k) = \text{`win'}\\\text{ and } 
            \mathcal{J}(a_k, a_j) = \text{`lose'}
            \end{aligned} \\[1em]
            v_j \rightarrow v_k, & 
            \begin{aligned}[t]
            \text{if } \mathcal{J}(a_j, a_k) = \text{`lose'}\\\text{ and } 
            \mathcal{J}(a_k, a_j) = \text{`win'}
            \end{aligned} \\[1em]
            v_j \leftrightarrow v_k, & \text{otherwise}
        \end{cases}
    \end{aligned}
\label{tournament}
\end{equation}
Here, $\mathcal{J}(a_j, a_k) \in \{\text{`win'}, \text{`lose'}\}$ denotes the pairwise comparison result between answers $a_j$ and $a_k$, where $a_j$ appears before $a_k$ in the prompt.

Considering the common position bias in evaluator LLMs~\cite{wang-etal-2024-large-language-models-fair}, we apply position swapping by comparing each response pair in both orders: $\mathcal{J}(a_j, a_k)$ and $\mathcal{J}(a_k, a_j)$. This ensures a more robust and balanced assessment~\cite{zheng2024judging_mt}. If the preferences differ across orders—indicating possible position bias—a bidirectional edge is added between the corresponding vertices to represent a `tie'.

\subsection{Quality Analysis Framework for Evaluator LLM Training Data}
\label{subsec:quality_analysis_criteria}
We introduce a framework for analyzing evaluator LLM training data quality. Leveraging directed graph representations of preference tournaments, the framework introduces two metrics: SCC analysis quantifying preference non-transitivity and two-dimensional structural entropy measuring preference clarity.

\textbf{Quantifying preference non-transitivity via SCC analysis.} 
SCCs are maximal subgraphs where any two vertices are mutually reachable through directed paths. This property provides a natural mechanism for identifying preference cycles—a direct manifestation of non-transitivity in evaluator LLM judgments. When vertex $v_i$ can reach vertex $v_j$ via a directed path, this indicates a preference chain $a_j \succ \cdot \succ a_i$. The simultaneous existence of paths from $v_i$ to $v_j$ and from $v_j$ to $v_i$ represents contradictory preference relationships, signaling a violation of transitivity.

We employ Tarjan's algorithm~\cite{doi:10.1137/0201010} to efficiently identify SCCs in our directed graphs. Tarjan's algorithm operates in linear time, specifically $\mathcal{O}(|V| + |E|)$, where $|V|$ and $|E|$ denote the number of vertices and edges, respectively, ensuring scalability to large graphs. Since non-transitivity requires at least three elements in a cycle, we focus on SCCs containing more than two vertices. Additionally, we exclude cases where every vertex pair shares bidirectional edges (indicating `tie'), as these represent consistent preference relations. For example, the pattern $(A = B)$, $(B = C)$, $(C = A)$ constitutes a valid SCC but maintains transitive preference relationships. Formally, we identify the set of non-transitive SCCs as:

\begin{equation}
\label{non-trans-scc}
\mathcal{S}_\text{n-t} = \left\{ S \in \text{SCCs}(G) \,\middle|\,
\begin{aligned}
|S| > 2 \land\exists v_j, v_k \in S, \\(v_j \leftrightarrow v_k) \notin E(S)
\end{aligned}
\right\},
\end{equation}

\noindent where $\text{SCCs}(G)$ represents all SCC in graph $G$, $|S|$ denotes the component size, and $(v_j \leftrightarrow v_k) \notin E(S)$ indicates the absence of bidirectional edges between some vertices.

To quantify the prevalence of non-transitivity across our entire training set, we compute the non-transitivity ratio:

\begin{equation}
\rho_{\text{non-trans}} = \frac{\sum_{q_i \in \mathbf{Q}} |S_{\text{n-t}}(G_i)|}{\sum_{q_i \in \mathbf{Q}} |V_i|},
\label{ep:measure_transitivity}
\end{equation}

\noindent where the numerator represents the total number of vertices in non-transitive SCCs across all questions $\mathbf{Q}$, and the denominator represents the total number of vertices across all graphs. This metric ranges from 0 to 1, with higher values indicating greater prevalence of non-transitive in the evaluator LLM's preferences. A perfectly transitive preference system would yield $\rho_{\text{non-trans}} = 0$, while completely cyclical preferences would approach $\rho_{\text{non-trans}} = 1$.

\textbf{Analysis of preference clarity based the structural entropy of directed graph.} While the SCC-based approach effectively quantifies preference non-transitivity, this metric alone is insufficient to characterize preference linearity. For instance, a dataset comprised entirely of preference `tie’ would exhibit perfect transitivity yet fail to establish any linear ordering. To address this limitation, we introduce directed graph entropy as a complementary measurement. Structural entropy~\cite{7456290} extends Shannon entropy~\cite{6773024} to directed graphs, providing a measure of system uncertainty and the complexity of relationships within the graph. These concepts have been widely applied across various domains
~\cite{10.1145/3616855.3635820,Duan_Chen_Liu_Liu_Yue_Li_2024,10.1145/3660522,Hou_Zhu_Liu_Su_Xia_Wu_Xu_2025}. We introduce and refine the two-dimensional structural entropy metric for directed graphs to analyze the overall clarity of preferences in evaluator LLMs.

A crucial observation drives our methodology is that SCCs composed of single vertices inherently exhibit strict transitivity. Preference relations among such singleton components naturally form a linear order that does not increase preference complexity. For example, consider preferences $A \succ B$, $B \succ C$, and $A \succ C$. These form a clear linear order $A \succ B \succ C$, where each vertex constitutes its own singleton SCC. This clarity reflects the inherent transitivity of singleton SCCs.
Based on this insight, we adopt SCCs as the basic community units when calculating structural entropy. To ensure accurate quantification of meaningful structural complexity, we exclude interactions between pure single-point SCCs. Instead, we retain only interactions between singleton SCCs and multi-vertex SCCs, as well as interactions between different multi-vertex SCCs. This selective approach ensures more accurate quantification of meaningful structural complexity. Consequently, preference relations closer to a linear order produce lower entropy values.
\begin{figure}[htbp]
  \centering
\includegraphics[width=1\linewidth]{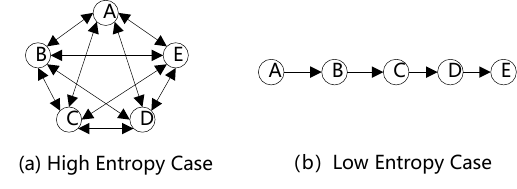}

\caption{Cases of High and Low Structural Entropy in Preference Tournaments.}
  \label{fig:cases} 
\end{figure}

Figure~\ref{fig:cases} illustrates two contrasting scenarios that provide an intuitive explanation of the two-dimensional structural entropy in tournament graph preference relations. Figure~\ref{fig:cases} (a) depicts a high entropy case with a chaotic tournament containing multiple preference cycles resulting in complex non-transitive relationships. In contrast, Figure~\ref{fig:cases} (b) shows a low entropy case with a perfectly ordered tournament where responses form a clear linear hierarchy, producing a simple transitive structure.

Given a directed graph $G = (V, E)$ with $n = |V|$ vertices, we perform SCC decomposition resulting in $\text{SCCs}(G) = \{SCC_1, SCC_2, \ldots, SCC_L\}$, where $L$ is the total number of SCCs. For any vertex $v \in V$, we denote $d_{\text{in}}(v)$ and $d_{\text{out}}(v)$ as the in-degree and out-degree of $v$ in $G$. For any $SCC_i$, we define its volume as $v(SCC_i) = \sum_{v \in SCC_i} d_{\text{in}}(v)$, while $v(G)$ represents the total in-degree of the entire graph.

The two-dimensional structural entropy is computed as:
{\small
\begin{align}
H^2(G) &= -\sum_{j=1}^{L} \frac{g_j}{v(G)} \log_2 \frac{v(SCC_j)}{v(G)} \nonumber \\
&- \sum_{j=1}^{L} \frac{v(SCC_j)}{v(G)} \left(\sum_{v \in SCC_j} \frac{d_\text{in}(v)}{v(SCC_j)} \log_2 \frac{d_\text{in}(v)}{v(SCC_j)}\right),
\end{align}
}

\noindent The first term captures the entropy of the partition (inter-community complexity), while the second term captures the weighted average of entropies within each SCC (intra-community complexity). The variable $g_j$ represents the number of incoming edges to $SCC_j$ from vertices outside this component, specifically counting edges between singleton SCCs and multi-vertex SCCs, as well as edges between different multi-vertex SCCs. This value quantifies the external influence on each SCC and plays a crucial role in measuring cross-community complexity.

To facilitate meaningful comparisons across graphs of different sizes, we normalize the structural entropy. The normalized structural entropy for a graph $G$ is defined as:

\begin{equation}
\tau(G) = \frac{H^2(G)}{\log_2 n}.
\end{equation}

Normalization is necessary since raw entropy increases with graph size. Dividing by $\log_2 n$ scales the entropy to a range between 0 and 1, where values closer to 0 indicate more ordered structures approaching linear hierarchy, while values closer to 1 suggest higher complexity and less clear preference ordering. The upper bound $\log_2 n$ represents the maximum possible entropy for a graph with $n$ vertices, corresponding to a completely uniform distribution of structural information. To evaluate the overall preference clarity of an evaluator LLM across a set of questions $\mathbf{Q}$, we compute the average normalized structural entropy:

\begin{equation}
\tau_{\text{avg}} = \frac{\sum_{q_i \in \mathbf{Q}} \tau(G_i)}{|\mathbf{Q}|},
\end{equation}
Here, $\tau(G_i)$ is the normalized structural entropy for the graph $G_i$ corresponding to question $q_i$, and $|\mathbf{Q}|$ is the total number of questions in the set $\mathbf{Q}$.

The $\tau_{\text{avg}}$ metric quantifies preference clarity: higher values indicate complex, non-transitive patterns, while lower values reflect relationships closer to linear order. It enables objective assessment of consistency across evaluator LLMs, where lower entropy denotes more coherent and interpretable preferences.
\subsection{Filtering Strategy for Preference Data That Induce Non-Transitivity}
\label{subsec:data_filtering}
If cycles within each SCC of a directed graph can be eliminated to form DAGs, the entire graph can be transformed into a DAG. Our framework leverages this property by converting each SCC into a DAG while preserving inter-SCC preference relations, ensuring global acyclicity without altering component relationships. From the resulting DAG, we derive transitive preference relations to filter evaluator LLMs' preference data, enabling training data self-purification. The procedure is:
\begin{enumerate}
  \item For each \(SCC_i\) compute the in-degree $ e_i^{\text{in}} $ for each vertex $ v_i $, representing its global `win' score. Vertices with higher in-degree scores are prioritized over those with lower scores. To reconstruct the internal edges within \(SCC_i\), first remove all original edges between vertices within \(SCC_i\). Then, for any pair \(v_i, v_j\): if \(e_i^{\text{in}} > e_j^{\text{in}}\), add a directed edge \(v_j \rightarrow v_i\); if \(e_i^{\text{in}} = e_j^{\text{in}}\), add a bidirectional edge \(v_i \leftrightarrow v_j\).
  \item After processing all SCCs, the original cyclic graph is transformed into a DAG. This DAG is used to filter the training data as follows: For bidirectional edges $(v_i \leftrightarrow v_j)$, the correct preferences are recorded as $\mathcal{J}(a_i, a_j) = \text{`tie'}$ and $\mathcal{J}(a_j, a_i) = \text{`tie'}$. For unidirectional edges $(v_i \rightarrow v_j)$, the correct preferences are recorded as $\mathcal{J}(a_i, a_j) = \text{`lose'}$ and $\mathcal{J}(a_j, a_i) = \text{`win'}$.
  \item The training data are traversed sequentially. All instances that are consistent with the correct preferences are added to the training set ``Cleaned", while the remaining data are placed in the training set ``Discarded".
\end{enumerate}
The final ``Cleaned" training data only retains linearly transitive preference data, resulting in $\rho_{\text{non-trans}} = 0$ and $\tau_{\text{avg}} = 0$.  Our approach balances optimality and scalability, with an overall time complexity of $\mathcal{O}(|V|^2 + |E|)$, consisting of $\mathcal{O}(|V|^2)$ for graph construction, $\mathcal{O}(|V| + |E|)$ for SCC decomposition, $\mathcal{O}(|V|^2)$ for intra-SCC reconstruction, and $\mathcal{O}(|E|)$ for edge filtering. The detailed algorithmic procedure is provided in Appendix A.

\section{Experiment Setup}
\begin{table*}
\centering
\setlength{\tabcolsep}{1mm}
\small
\begin{tabular}{lccccccccccccccc}\toprule
\multicolumn{1}{l}{\multirow{2}*{{\textbf{Model}}}} &\multicolumn{2}{c}{{\textbf{Helpful\_Base}}}  &\multicolumn{2}{c}{{\textbf{Vicuna}}} &\multicolumn{2}{c}{{\textbf{Oasst}}} &\multicolumn{2}{c}{{\textbf{Koala}}}&\multicolumn{2}{c}{{\textbf{Self-instruct}}}\\
\cmidrule(r){2-3}
\cmidrule(r){4-5}%
\cmidrule(r){6-7}%
\cmidrule(r){8-9}%
\cmidrule(r){10-11}%
\multicolumn{1}{c}{~} & $\rho_{\text{non-trans}} ^\downarrow$ &$\tau_{\text{avg}}^\downarrow$ & $\rho_{\text{non-trans}} ^\downarrow$  &$\tau_{\text{avg}}^\downarrow$& $\rho_{\text{non-trans}} ^\downarrow$ &$\tau_{\text{avg}}^\downarrow$& $\rho_{\text{non-trans}} ^\downarrow$ &$\tau_{\text{avg}}^\downarrow$& $\rho_{\text{non-trans}} ^\downarrow$  &$\tau_{\text{avg}}^\downarrow$
\\ 
\midrule
\textbf{Stronger LLMs}\\
\midrule
Qwen2.5-Max   & 63.7\%  & 0.805 &75.4\%  &0.845 &64.3\%&0.788&71.5\%&0.830&65.0\%&0.780\\  
\midrule
\textbf{Base models and Variants}\\
\midrule
Qwen-Base  & 82.8\%  & 0.922&  78.9\%  &0.891 &83.4\%  &0.914& 81.0\% & 0.910& 81.5\% &0.912 \\ 
Qwen-Raw & 62.0\% & 0.796 &57.5\% &0.803&55.8\%&0.773&64.3\%&0.816&59.7\%&0.767\\
Qwen-Random &60.6\%&0.790&63.6\%&0.808&58.8\%&0.771&60.4\%&0.812&57.1\%&0.764 \\
\textbf{Qwen-Cleaned (ours)}  &\textbf{44.9\%}&\textbf{0.700}&\textbf{43.9\%}&\textbf{0.726}&\textbf{47.0\%}&\textbf{0.694}&\textbf{48.5\%}&\textbf{0.715}&\textbf{49.0\%}&\textbf{0.680}\\

LLaMA-Base  &76.4\%&0.852&67.0\%&0.846&67.4\%&0.793&69.9\%&0.818&71.0\%&0.809\\ 
LLaMA-Raw   &59.0\%&0.765&60.2\%&0.772&58.2\%&0.760&57.2\%&0.746&60.0\%&0.783\\ 
\textbf{LLaMA-Cleaned (ours)}  &\textbf{40.2\%}&\textbf{0.652}&\textbf{45.4\%}&\textbf{0.691}&\textbf{43.0\%}&\textbf{0.629}&\textbf{44.4\%}&\textbf{0.659}&\textbf{42.0\%}&\textbf{0.642}\\
\bottomrule
\end{tabular}
\caption{\label{tab:main experiment} 
Comparison of \textbf{Preference Non-Transitivity} and \textbf{Overall Clarity} for evaluator LLMs. 
Qwen-Base denotes to the original Qwen2.5-7B-Instruct model, while Qwen-Raw, Qwen-Random, and Qwen-Cleaned denote models fine-tuned on the \textbf{``Raw"}, \textbf{``Random"}, and \textbf{``Cleaned"} training sets, respectively. For example, Qwen-Cleaned in the Helpful\_Base column reflects the performance of the model fine-tuned on the "Cleaned" training set derived from filtered Helpful\_Base training set.
}
\end{table*}

\subsection{Dataset}
In this study, we conduct experimental validation using the AlpacaEval benchmark~\cite{alpaca_eval}. AlpacaEval is specifically designed to assess the overall capabilities of LLMs in open-ended tasks, covering a wide range of evaluation scenarios such as reasoning and text and code generation. The benchmark comprises five datasets, Helpful\_Base, Oasst, Koala, Vicuna, and Self-Instruct.

\subsection{Preference Data Collection}
We evaluated Qwen2.5-Max~\cite{qwen25max} on 2.5k human-annotated samples from AlpacaEval using the chain-of-thought (COT) comparison~\cite{alpaca_eval}. Qwen2.5-Max achieved a human agreement rate of 68.9\%. Based on these results, we selected it as our teacher model for preference data generation.

\subsection{Experiment Details}
\begin{figure}[htbp]
  \includegraphics[width=\columnwidth]{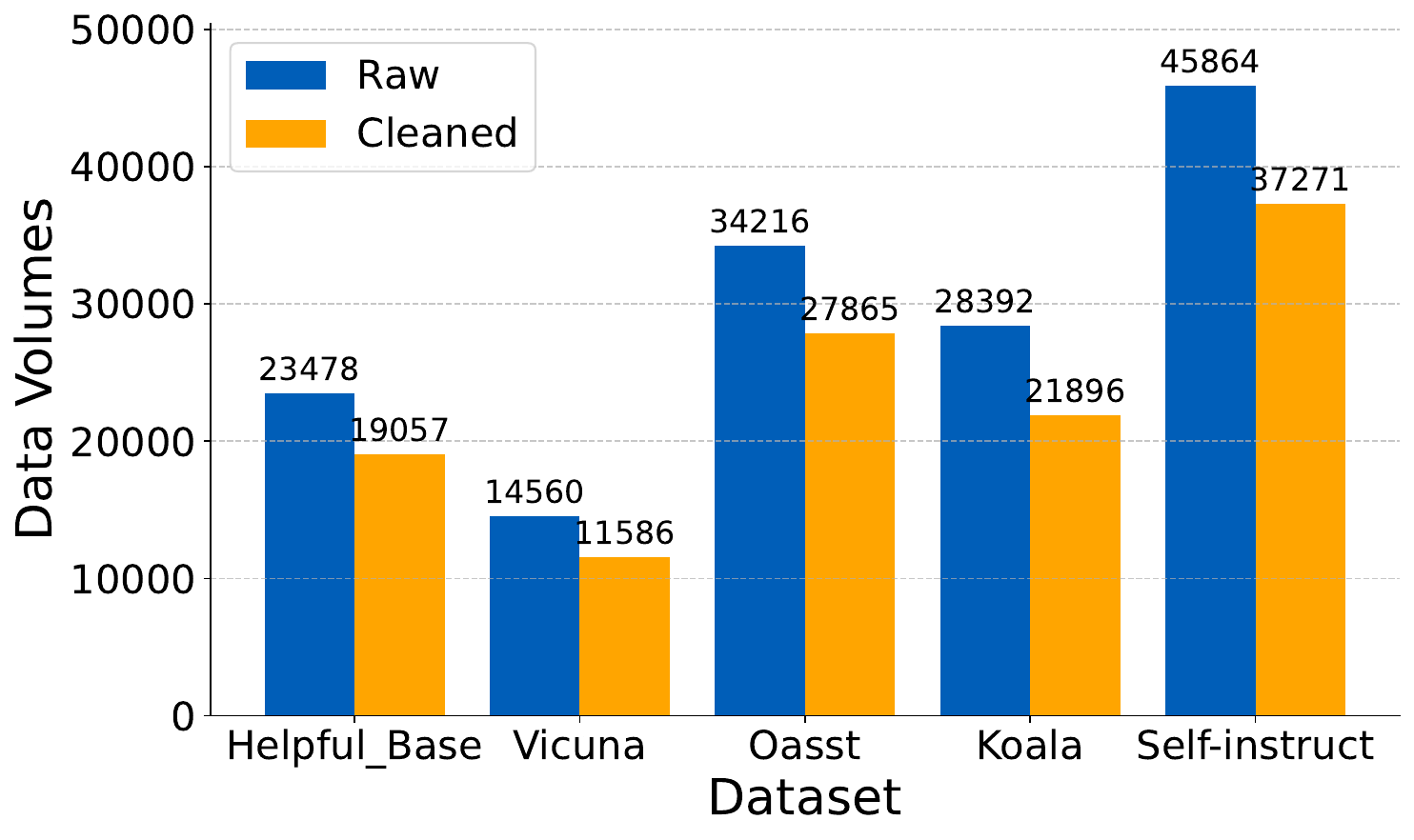}
\caption{\label{fig:Data_Volumes_comparison}Comparison of data volumes between \textbf{``Raw"} and \textbf{``Cleaned"} training sets across datasets. The \textbf{``Cleaned"} training set's volume is approximately 80\% of the \textbf{``Raw"} training set for each dataset.}
\end{figure}

21 representative LLMs were selected from the AlpacaEval leaderboard (14 for training, 7 for testing). All experiments employed CoT comparison templates with temperature set to 0, and preferences were generated via the Qwen2.5-Max API. The unfiltered ``Raw'' datasets were processed according to the procedure described in Section~\ref{subsec:data_filtering} to produce ``Cleaned'' training sets; the corresponding distributions are illustrated in Figure~\ref{fig:Data_Volumes_comparison}. Qwen2.5-7B-Instruct was fine-tuned on both versions of each of the five datasets using LoRA~\cite{hu2022lora} (rank~$=8$, 3~epochs, learning rate~$=1 \times 10^{-4}$, batch size~$=16$). 

\section{Results and Analysis}

This section verifies the effectiveness of the proposed filtering method on five datasets. The analysis begins with an examination of preference non-transitivity degree and overall preference clarity of evaluator LLMs. To further demonstrate the method's effectiveness, experiments based on actual usage scenarios are conducted using the MT-bench evaluation framework to calculate the standard deviation between final results. Additionally, a detailed data quality analysis examines two key aspects of manual evaluation: (1) consistency between manual annotators, and (2) consistency between evaluation results and human majority vote.
\subsection{Main Results}
\textbf{Analysis of preference non-transitivity and clarity.} As shown in Table~\ref{tab:main experiment}, models fine-tuned on the ``Cleaned" training set demonstrate the lowest preference non-transitivity and highest preference clarity across all datasets, including Qwen2.5-Max, highlighting the effectiveness of our proposed data filtering methodology. To verify that models fine-tuned on the ``Cleaned" training set achieve lower preference non-transitivity and higher preference clarity even on ``unseen" questions, we conducted cross-validation as our out-of-domain experiment by testing each fine-tuned model across five test sets. As demonstrated in Table~\ref{tab:qwen_nontrans}, models fine-tuned on ``Cleaned" training set exhibited an average reduction in non-transitivity of 13.8\% compared to those trained on ``Raw" training set. Table~\ref{tab:qwen_clarity} shows a reduction of 0.088 in normalized structural entropy, indicating clearer overall preferences. Detailed results are in Appendix B.

\begin{table}[t]
\centering
\small
\begin{tabular}{l|ccc}
\hline
\multicolumn{1}{c|}{\textbf{Dataset}} & \textbf{Raw} & \textbf{Cleaned} & \textbf{$\Delta$} \\ \hline
\textbf{Helpful\_Base}  & 66.0\% & 50.3\% & -15.7\% \\
\textbf{Vicuna}        & 61.6\% & 51.0\% & -10.6\% \\
\textbf{Oasst}         & 62.6\% & 48.7\% & -13.9\% \\
\textbf{Koala}         & 64.1\% & 48.7\% & -15.4\% \\
\textbf{Self-instruct} & 67.2\% & 53.8\% & -13.4\% \\ \hline
\textbf{Average}       & 64.3\% & 50.5\% & -13.8\% \\ \hline
\end{tabular}
\caption{\label{tab:qwen_nontrans}Comparison of $\rho_{\text{non-trans}}$ (\textbf{average preference non-transitivity}) between models fine-tuned on \textbf{``Raw"} and \textbf{``Cleaned"} training sets using Qwen2.5-7B-Instruct.}
\end{table}

\begin{table}[t]
\centering
\small
\begin{tabular}{l|ccc}
\hline
\multicolumn{1}{c|}{\textbf{Dataset}} & \textbf{Raw} & \textbf{Cleaned} & \textbf{$\Delta$} \\ \hline
\textbf{Helpful\_Base}  & 0.820 & 0.717 & -0.103 \\
\textbf{Vicuna}        & 0.806 & 0.737 & -0.069 \\
\textbf{Oasst}         & 0.797 & 0.710 & -0.087 \\
\textbf{Koala}         & 0.804 & 0.718 & -0.086 \\
\textbf{Self-instruct} & 0.829 & 0.735 & -0.094 \\ \hline
\textbf{Average}       & 0.811 & 0.723 & -0.088 \\ \hline
\end{tabular}
\caption{\label{tab:qwen_clarity}Comparison of $\tau_{\text{avg}}$ (\textbf{average preference clarity}) between models fine-tuned on \textbf{``Raw"} and \textbf{``Cleaned"} training sets using Qwen2.5-7B-Instruct.}
\end{table}

\textbf{Human validation of discarded data quality.} To verify the low quality of discarded data, we evaluated 100 randomly sampled instances from both ``Cleaned" and ``Discarded" sets across five datasets (1,000 total instances). Three independent annotators assessed these samples, enabling measurement of inter-annotator consistency and model-human alignment. Results in Table~\ref{tab:compact_evaluation} indicate that human annotator consistency in the ``Discard" training set averages only 34.4\%, significantly lower than the 52.6\% observed in the ``Cleaned" training set. This confirms that data leading to non-transitive preferences is inherently ambiguous. Furthermore, model evaluation reveals that consistency between the ``Discarded" training set and human majority votes averages merely 51.2\%, substantially below the 80.6\% achieved by the ``Cleaned" training set. These findings provide compelling evidence that the ``Discarded" training set contains low-quality preference pairs unsuitable for training models to learn consensus human preferences.

For deeper insight into non-transitive preferences, we observed that source response pairs exhibiting non-transitivity demonstrate higher textual similarity. This observation suggests that preference cycles are likely to occur when the quality difference between source responses falls below the just noticeable difference (JND)~\cite{stern2010just} threshold of the evaluator LLM. As shown in Table \ref{tab:self_bleu}, we quantified content similarity between response pairs by calculating Self-Bleu scores. Source response pairs with non-transitive preference relations consistently displayed higher Self-Bleu values than those with transitive relations, confirming that preference non-transitivity correlates with greater text similarity between comparative samples.

\begin{table}[t]
\small
\centering
\begin{tabular}{lcccc}
\toprule
& \multicolumn{2}{c}{\textbf{HC}} & \multicolumn{2}{c}{\textbf{MHA}} \\
\cmidrule(lr){2-3} \cmidrule(lr){4-5}
\textbf{Dataset} & \textbf{Cleaned} & \textbf{Discarded} & \textbf{Cleaned} & \textbf{Discarded} \\
\midrule
\textbf{Help.} & 59\% & 46\% & 72\% & 45\% \\
\textbf{Vicuna} & 48\% & 24\% & 78\% & 47\% \\
\textbf{Oasst} & 56\% & 33\% & 83\% & 53\% \\
\textbf{Koala} & 54\% & 41\% & 84\% & 61\% \\
\textbf{Self.} & 46\% & 28\% & 86\% & 50\% \\\hline
\textbf{Average}  & 52.6\% & 34.4\% & 80.6\% & 51.2\% \\ \hline
\end{tabular}
\caption{\label{tab:compact_evaluation}Comparison of Human Evaluation Consistency and Model-Human Agreement between \textbf{training sets}. HC: Human Consistency (inter-annotator consensus); MHA: Model-Human Agreement (alignment with majority vote). Help. = Helpful\_Base; Self. = Self-instruct.}
\end{table}

\begin{table}[t]
\small
\footnotesize
\centering
\begin{tabular}{l ccccc}
\toprule
\textbf{Pref.} & \textbf{Helpful.} & \textbf{Vicuna} & \textbf{Oasst} & \textbf{Koala} & \textbf{Self.} \\
\midrule
\textbf{Non.} & 0.0903 & 0.1026 & 0.0946 & 0.0888 & 0.1135 \\
\textbf{Trans.} & 0.0713 & 0.0918 & 0.0788 & 0.0746 & 0.0829 \\
\bottomrule
\end{tabular}
\caption{Comparison of \textbf{source text pair} Self-Bleu scores across datasets. Non. and Trans. indicate source pairs from non-transitive and transitive preferences; Helpful. = Helpful\_Base; Self. = Self-instruct.}
\label{tab:self_bleu}
\end{table}
\textbf{Performance analysis of real-world evaluation systems.} We compared the performance of our trained evaluator LLMs within the MT-bench evaluation framework~\cite{zheng2024judging_mt}. We adopted the adjusted win rate calculation consistent with the MT-bench evaluation methodology. This metric normalizes performance by treating ties as partial wins, defined as $r_{\text{adj}} = (r_w + 0.5 \cdot r_t)/(r_w + r_l + r_t)$, where $r_w$, $r_l$, and $r_t$ represent the win rate, loss rate, and tie rate respectively. The adjusted rate $r_{\text{adj}}$ serves as the primary indicator for our model ranking. Results in Table~\ref{tab:std_analysis} demonstrate that models fine-tuned on the ``Cleaned" training sets exhibit significantly higher standard deviations in their adjusted win rates across five datasets. This indicates enhanced robustness of the evaluation results and better differentiation of performance differences between models. 
\begin{table}[htbp]
\small
\centering
\begin{tabular}{l ccccc}
\toprule
\textbf{Dataset} & \textbf{Help.} & \textbf{Vicu.} & \textbf{Oasst} & \textbf{Koala} & \textbf{Self.} \\
\midrule
Cleaned & 16.1\% & 16.6\% & 13.1\% & 14.9\% & 9.7\% \\
Raw     & 12.2\% & 13.6\% & 11.0\% & 12.3\% & 8.2\% \\
\midrule
$\Delta$ 
        & +3.9\% & +3.0\% & +2.1\% & +2.6\% & +1.5\% \\
\bottomrule
\end{tabular}
\caption{\label{tab:std_analysis}Standard deviation analysis. Help. = Helpful\_Base; Vicu. = Vicuna; Self. = Self-instruct.}
\end{table}
\begin{table}[t]
\small
\centering
\begin{tabular}{lcccc}
\toprule
& \multicolumn{2}{c}{\textbf{MHA}} & \multicolumn{2}{c}{\textbf{SC}} \\
\cmidrule(lr){2-3} \cmidrule(lr){4-5}
\textbf{Dataset} & \textbf{Raw} & \textbf{Cleaned} & \textbf{Raw} & \textbf{Cleaned} \\
\midrule
\textbf{Helpful\_base} & 66.9\% & 66.9\% & 0.93 & 0.97 \\
\textbf{Vicuna} & 65.2\% & 65.8\% & 0.97 & 0.98 \\
\textbf{Oasst} & 66.4\% & 67.6\% & 0.93 & 0.93 \\
\textbf{Koala} & 66.4\% & 66.9\% & 0.98 & 0.98 \\
\textbf{Self-instruct} & 67.8\% & 68.3\% & 0.98 & 1.00\\\hline
\textbf{Average}  & 66.5\% & 67.1\% & 0.96 & 0.97 \\ \hline
\end{tabular}
\caption{\label{tab:qwen_ave_human}Comparison of Model-Human Agreement (MHA) and Spearman Correlation (SC) between \textbf{fine-tuned models}. Spearman Correlation quantifies the rank correlation between model and human preferences.}
\end{table}

\textbf{Analysis of human agreement impact.} The model's alignment with human preferences was evaluated by calculating agreement rates and Spearman correlation coefficients on 2.5k manually annotated preference labels from AlpacaEval. Results in Table~\ref{tab:qwen_ave_human} indicate that models fine-tuned on the ``Cleaned" training set consistently outperformed those trained on ``Raw" training set across multiple dataset, with maximum improvements of 1.2\% in human agreement and 0.04 in Spearman correlation. These results demonstrate that filtering ambiguous, low-quality preference data not only improves discrimination and enhances the performance of downstream evaluation systems but also increases consistency with human judgment, providing strong evidence for practical applications.
\subsection{Ablation Studies}
\textbf{Effect of Different Base Models:} Testing using the same steps with LLaMA3.1-8B-Instruct confirms our findings: models trained on the ``Cleaned" training set consistently outperformed those trained on the ``Raw" training set (Table~\ref{tab:main experiment}). \textbf{Data Filtering Analysis:} Our ``Cleaned" set retained approximately 80\% of the raw data (Figure~\ref{fig:Data_Volumes_comparison}). A control experiment with random 20\% filtering (Qwen-Random) produced results similar to training on the ``Raw" set but with higher preference non-transitivity, confirming that our targeted approach improves data quality beyond simple reduction (Table~\ref{tab:main experiment}). \textbf{Prompt Format Variations:} Additional experiments with prompts explicitly allowing ``tie" judgments further validated our methodology. Detailed prompt templates and results are provided in Appendices B and C.

\section{Conclusion}
We propose ELSPR, a graph-theoretic framework to analyze and mitigate non-transitivity in evaluator LLMs by modeling pairwise comparisons as a tournament graph and filtering problematic data. Ambiguous, low-quality preferences are a major source of non-transitive judgments, and targeted filtering improves evaluation reliability. Empirical results show that filtered data aligns better with human judgments and can fine-tune models to reduce non-transitive bias and structural entropy. This work underscores the importance of data quality and provides a scalable approach to enhance consistency on future benchmarks.
\section*{Acknowledgments}
This work is supported by the National Natural Science Foundation of China under Grant Nos. 62572105 and U22B2005; the Liaoning Revitalization Talents Program under Grant No. XLYC2403086.
\bibliography{aaai2026}

\newpage 
\appendix
\section{Algorithm Detail}
\label{sec:filter_detail}
In this section, we provide the detailed pseudocode for the filtering algorithm described in Section 3.3 of the main text.
\begin{algorithm}[htbp]
\caption{Filtering Strategy for Preference Data That Induces Non-Transitivity}
\label{alg:data_filtering}
\begin{algorithmic}[1]
\Require Cyclic directed graph \( G = (V, E) \)
\Ensure Clean dataset with transitive preference relations

\State Decompose \( G \) into strongly connected components \(SCCs\): \( \{SCC_1, SCC_2, \dots, SCC_n\} \)
\For{each \( SCC_i \in \{SCC_1, \dots, SCC_n\} \)}
    \For{each vertex \( v_k \in SCC_i \)}
        \State Compute in-degree score \( e_k^{\text{in}} \) 
    \EndFor
     \State Remove edges within \( SCC_i \) 
    \For{each pair of vertices \( (v_i, v_j) \in SCC_i \)}
        \If{ \( e_i^{\text{in}} > e_j^{\text{in}} \) }
            \State Add edge \( (v_j \rightarrow v_i) \)
        \ElsIf{ \( e_i^{\text{in}} = e_j^{\text{in}} \) }
            \State Add bidirectional edge \( (v_i \leftrightarrow v_j) \)
        \EndIf
    \EndFor
\EndFor
\State Combine all modified \( SCC_i \) components to form a global DAG \( G' \)
\For{each edge in \( G' \)}
    \If{edge is \( (v_i \rightarrow v_j) \)}
        \State Set \( \mathcal{J}(a_i, a_j) = \text{`lose'} \), \( \mathcal{J}(a_j, a_i) = \text{`win'} \)
    \ElsIf{edge is \( (v_i \leftrightarrow v_j) \)}
        \State Set \( \mathcal{J}(a_i, a_j) = \mathcal{J}(a_j, a_i) = \text{`tie'} \)
    \EndIf
\EndFor

\State Initialize empty dataset \texttt{Cleaned}
\State Initialize empty dataset \texttt{Discarded}
\For{each data point in original dataset}
    \If{preference relation matches \( \mathcal{J} \)}
        \State Add data point to \texttt{Cleaned}
    \Else
        \State Add data point to \texttt{Discarded}
    \EndIf
\EndFor
\State \Return \texttt{Cleaned}, \texttt{Discarded}
\end{algorithmic}
\end{algorithm}
\section{Additional Experimental Results}

\subsection{Experimental Results of Different Prompt Forms.}
\label{sec:allow_tie_result}
\begin{figure}[htbp]
  \includegraphics[width=1.0\columnwidth]{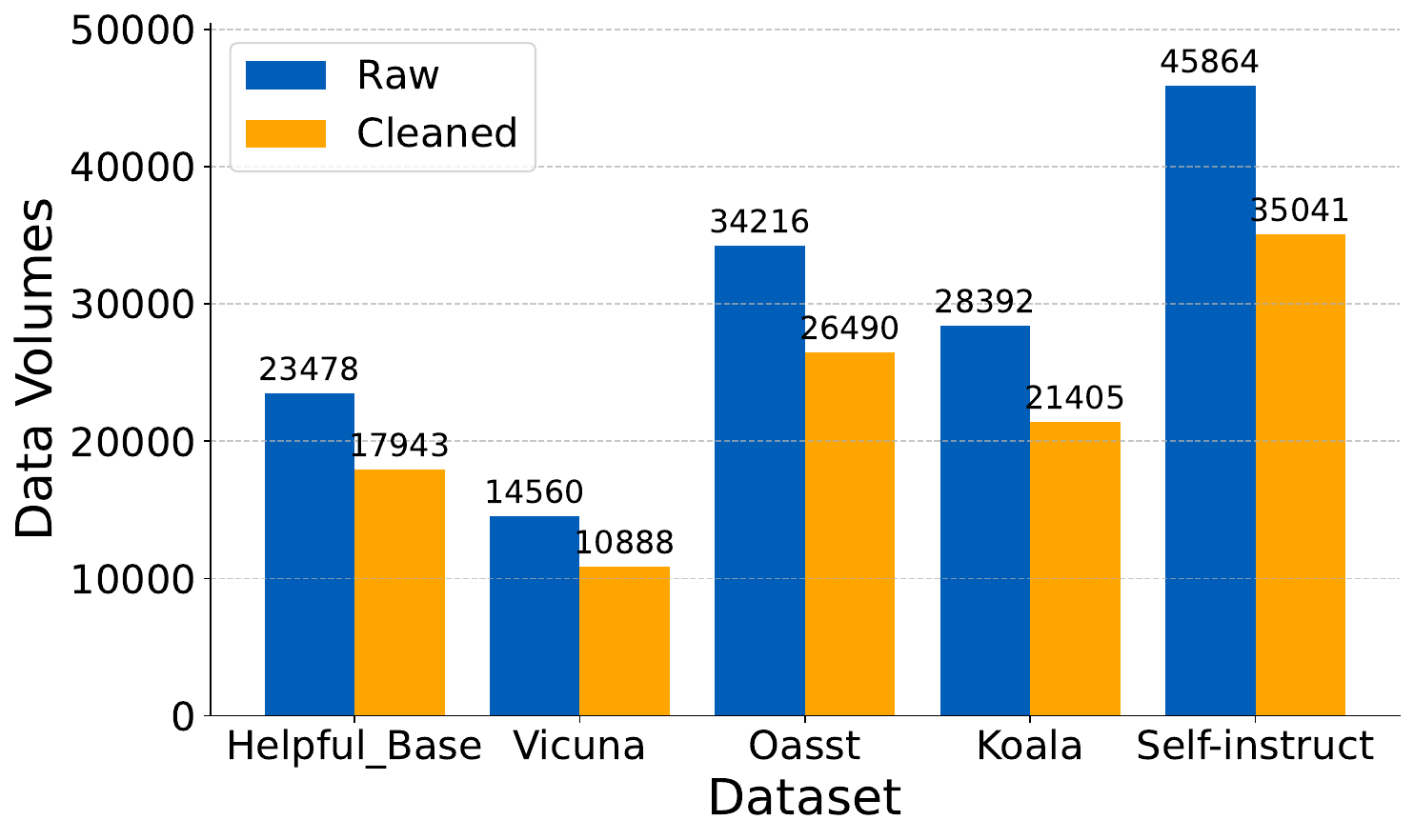}  \caption{Comparison of Data Volumes Between ``Raw" and ``Cleaned" Training Sets Across Different Datasets (Using the CoT Comparison (Tie Allowed) Prompt Template)}
  \label{Data_Volumes_comparison_2}
\end{figure}

We repeated the experiment using the CoT Comparison (Tie Allowed) prompt template.
Figure~\ref{Data_Volumes_comparison_2} illustrates the distribution of dataset sizes before and after filtering across different training sets. The test results are presented in Table~\ref{allow_tie_result}.

Across five testing sets, models fine-tuned on the ``Cleaned" training set exhibit the highest preference clarity. In four of these testing sets (excluding Vicuna), these models also demonstrate the lowest preference non-transitivity, outperforming models fine-tuned on the ``Raw" training set and the teacher model Qwen2.5-Max. We observe that models fine-tuned on the ``Raw" training set, as well as the teacher model Qwen2.5-Max, exhibit a higher rate of ties in preferences on the Vicuna testing set. The prevalence of such ties—often arising from SCCs formed by cases like $(A = B)$, $(B = C)$, and $(C = A)$—significantly reduces the degree of preference non-transitivity when these SCCs are removed. As shown in Table~\ref{tab:vicuna_model_comparison}, we compare the probability of ties among three models on the Vicuna testing set. Furthermore, the $ \tau_{\text{avg}}^\downarrow $ values of models fine-tuned on the ``Cleaned" training set indicate a higher overall clarity in preferences.

\begin{table}[htbp]
\centering
\begin{tabular}{lc}
\hline
\textbf{Model} & tie probability  \\ \hline
Qwen2.5-Max & 7.59\% \\ 
Qwen-Raw    & 10.06\% \\ 
Qwen-Cleaned & 6.13\% \\ \hline
\end{tabular}
\caption{Comparison of tie probabilities in evaluation results produced by different evaluator LLMs on Vicuna testing set.}
\label{tab:vicuna_model_comparison}
\end{table}

\subsection{``Unseen" question validation}
\label{sec:cross-validation}
We perform a detailed comparison of the performance of models fine-tuned on different training sets and the advanced evaluator LLM on various testing sets:
\begin{itemize}
    \item Results for models based on Qwen2.5-7B-Instruct are presented in Table~\ref{tab:cross-qwen-result}.
    \item Results for models based on LLaMA3.1-8B-Instruct are presented in Table~\ref{tab:cross-LLaMA-result}.
\end{itemize}

\section{Evaluate Prompts}
\subsection{CoT Comparison}
\label{sec:prompt_template_cot}
The CoT Comparison prompt template, identical to AlpacaEval 2.0, is detailed in Table~\ref{tab:prompts-cot}.

\subsection{CoT Comparison (Tie Allowed)}
\label{sec:prompt_template_cot_tie_allowed}
The CoT Comparison (Tie Allowed) prompt template, is detailed in Table~\ref{tab:prompts-cot-allow-tie}.
\section{Limitations and Future Work}

Although we have validated the effectiveness of our filtering approach on five datasets from AlpacaEval, it still cannot fully cover all real-world scenarios, and further validation across a broader range of domains is needed. Additionally, while our method has successfully reduced the issue of preference non-transitivity, it has not been completely eliminated. Further efforts to minimize preference intransitivity remain an important direction for future work. Lastly, our study primarily focused on pairwise comparison tasks in single-turn dialogues, exploring multi-turn dialogue tasks presents an intriguing avenue for future research.

\section{Ethics Statement}
This research aims to analyze and mitigate non-transitive preferences in large language models when employed as evaluators for open-ended tasks. Our methodology exclusively utilizes publicly available data, existing datasets, and open-source frameworks for computational experiments. All human annotation revisions were conducted by graduate students under fair working conditions with appropriate compensation, ensuring both high-quality data collection and adherence to ethical standards. We confirm that our research procedures comply with established ethical guidelines for AI research and do not present foreseeable negative societal impacts.

\begin{table*}
\centering
\resizebox{\linewidth}{!}{ 
\begin{tabular}{lccccccccccccccc}\toprule
\multicolumn{1}{l}{\multirow{2}*{{\textbf{Model}}}} &\multicolumn{2}{c}{{\textbf{Helpful\_Base}}}  &\multicolumn{2}{c}{{\textbf{Vicuna}}} &\multicolumn{2}{c}{{\textbf{Oasst}}} &\multicolumn{2}{c}{{\textbf{Koala}}}&\multicolumn{2}{c}{{\textbf{Self-instruct}}}\\
\cmidrule(r){2-3}
\cmidrule(r){4-5}%
\cmidrule(r){6-7}%
\cmidrule(r){8-9}%
\cmidrule(r){10-11}%
\multicolumn{1}{c}{~} 
& $\rho_{\text{non-trans}}^\downarrow$ & $\tau_{\text{avg}}^\downarrow$ 
& $\rho_{\text{non-trans}}^\downarrow$ & $\tau_{\text{avg}}^\downarrow$
& $\rho_{\text{non-trans}}^\downarrow$ & $\tau_{\text{avg}}^\downarrow$
& $\rho_{\text{non-trans}}^\downarrow$ & $\tau_{\text{avg}}^\downarrow$
& $\rho_{\text{non-trans}}^\downarrow$ & $\tau_{\text{avg}}^\downarrow$
\\  
\midrule
Qwen2.5-Max   & 76.97\%  & 0.8654 
&\textbf{78.04\%}  &0.9231 
&79.71\%&0.8659
&76.31\%&0.8923
&71.77\%&0.8595\\ 
Qwen-Base  & 85.94\%  & 0.9371&  
90.71\%  &0.9602
&84.88\%  &0.9459
& 81.96\% & 0.9626
& 76.64\% &0.8853 \\ 
Qwen-Raw & 79.07\%  & 0.9244
&83.21\%  &0.9396
&77.58\%&0.9066
&78.66\%&0.9078
&73.87\%&0.8853\\
\textbf{Qwen-Cleaned (ours)} &\textbf{74.42\%}&\textbf{0.8596}
&85.89\%&\textbf{0.8922}
&\textbf{73.94\%}&\textbf{0.8564}
&\textbf{72.34\%}&\textbf{0.8403}
&\textbf{69.05\%}&\textbf{0.8288}\\ 
\bottomrule
\end{tabular}
}
\caption{\label{allow_tie_result} 
Comparison of \textbf{Preference Non-Transitivity} and \textbf{Overall Clarity} for evaluator LLMs ( using the CoT Comparison (Tie Allowed) prompt template). 
Qwen-Base denotes to the original Qwen2.5-7B-Instruct model, while Qwen-Raw and Qwen-Cleaned denote models fine-tuned on the \textbf{``Raw"} and \textbf{``Cleaned"} training sets, respectively. For example, Qwen-Cleaned in the Helpful\_Base column reflects the performance of the model fine-tuned on the ``Cleaned" training set derived from filtered Helpful\_Base data.
}
\end{table*}

\begin{table*}
\centering
\resizebox{\linewidth}{!}{ 
\begin{tabular}{lcccccccccccc}\toprule
\multicolumn{1}{l}{\multirow{2}{*}{\textbf{Model}}} 
& \multicolumn{2}{c}{\textbf{Helpful\_Base}}  
& \multicolumn{2}{c}{\textbf{Vicuna}} 
& \multicolumn{2}{c}{\textbf{Oasst}} 
& \multicolumn{2}{c}{\textbf{Koala}}
& \multicolumn{2}{c}{\textbf{Self-instruct}}
& \multicolumn{2}{c}{\textbf{Average}}\\
\cmidrule(r){2-3}
\cmidrule(r){4-5}%
\cmidrule(r){6-7}%
\cmidrule(r){8-9}%
\cmidrule(r){10-11}%
\cmidrule(r){12-13}%
\multicolumn{1}{c}{~} 
& $\rho_{\text{non-trans}}^\downarrow$ & $\tau_{\text{avg}}^\downarrow$ 
& $\rho_{\text{non-trans}}^\downarrow$ & $\tau_{\text{avg}}^\downarrow$
& $\rho_{\text{non-trans}}^\downarrow$ & $\tau_{\text{avg}}^\downarrow$
& $\rho_{\text{non-trans}}^\downarrow$ & $\tau_{\text{avg}}^\downarrow$
& $\rho_{\text{non-trans}}^\downarrow$ & $\tau_{\text{avg}}^\downarrow$
& $\rho_{\text{non-trans}}^\downarrow$ & $\tau_{\text{avg}}^\downarrow$
\\
\midrule
\textbf{Stronger LLMs}\\
\midrule
Qwen2.5-Max   & 63.68\%  & 0.8047 &75.36\%  &0.8448 &64.29\%&0.7883&71.52\%&0.8296&64.97\%&0.7803 &67.96\%&0.8095\\ 

\midrule
\textbf{Base Model}\\
\midrule
Qwen-Base  & 82.83\%  & 0.9221&  78.93\%  &0.8912 &83.36\%  &0.9138& 80.96\% & 0.9097& 81.46\% &0.9123 &81.51\% &0.9098\\ 
\midrule
\textbf{Cleaned group}\\
\midrule
Qwen-Helpful\_Base-Cleaned & 44.85\% & 0.6997 & 48.93\% & 0.7371 & 52.13\% & 0.7137 & 53.75\% & 0.7223 & 51.98\% & 0.7138&50.33\% &0.7173\\
Qwen-Vicuna-Cleaned & 47.18\% & 0.7395 & 43.93\% & 0.7262 & 52.43\% & 0.7276 & 53.48\% & 0.7418 & 57.77\% & 0.7483 &50.96\%&0.7367\\
Qwen-Oasst-Cleaned & 45.74\% & 0.7040 & 45.00\% & 0.7109 & 46.96\% & 0.6942 & 54.85\% & 0.7186 & 51.08\% & 0.7227 &48.73\%&0.7101\\
Qwen-Koala-Cleaned & 50.28\% & 0.7357 & 46.79\% & 0.7045 & 48.94\% & 0.7183 & 48.53\% & 0.7149 & 48.70\% & 0.7150 &48.65\%&0.7177\\
Qwen-Self-instruct-Cleaned & 56.26\% & 0.7504 & 51.79\% & 0.7489 & 54.94\% & 0.7370 & 57.23\% & 0.7590 & 48.98\% & 0.6802&53.84\%&0.7351 \\
\midrule
\textbf{Raw group}\\
\midrule
Qwen-Helpful\_Base-Raw & 62.02\% & 0.7963 & 74.29\% & 0.8626 & 63.07\% & 0.8109 & 69.87\% & 0.8385 & 60.49\% & 0.7936 &65.95\% &0.8204\\
Qwen-Vicuna-Raw & 61.13\% & 0.8147 & 57.50\% & 0.8034 & 64.59\% & 0.8019 & 60.35\% & 0.8150 & 64.23\% & 0.7966 &61.56\%&0.8063 \\
Qwen-Oasst-Raw & 60.58\% & 0.8099 & 70.89\% & 0.8101 & 55.78\% & 0.7734 & 63.83\% & 0.8150 & 61.96\% & 0.7746 &62.61\%&0.7966 \\
Qwen-Koala-Raw & 64.23\% & 0.8228 & 67.86\% & 0.8090 & 64.59\% & 0.7966 & 64.29\% & 0.8159 & 59.30\% & 0.7779 &64.05\%&0.8044 \\
Qwen-Self-instruct-Raw & 68.99\% & 0.8569 & 71.79\% & 0.8595 & 68.84\% & 0.8167 & 66.85\% & 0.8433 & 59.69\% & 0.7665&67.23\%&0.8286\\
\bottomrule
\end{tabular}
} 
\caption{\label{tab:cross-qwen-result}
Comparison of \textbf{Preference Non-Transitivity} and \textbf{Overall Clarity} for evaluator LLMs. 
Qwen-Base denotes to the original Qwen2.5-7B-Instruct model. \textbf{Raw group} and \textbf{Cleaned group} refer to models fine-tuned on the original and filtered training sets, respectively. 
Specifically, Qwen-Helpful\_Base-Raw is fine-tuned on the \textbf{``Raw"} training data generated from the \textbf{Helpful\_Base} dataset, while Qwen-Helpful\_Base-Cleaned is fine-tuned on the corresponding \textbf{``Cleaned"} dataset after removing non-transitive preference data.
}
\end{table*}

\begin{table*}
\centering
\resizebox{\linewidth}{!}{ 
\begin{tabular}{lcccccccccccc}\toprule
\multicolumn{1}{l}{\multirow{2}{*}{\textbf{Model}}} 
& \multicolumn{2}{c}{\textbf{Helpful\_Base}}  
& \multicolumn{2}{c}{\textbf{Vicuna}} 
& \multicolumn{2}{c}{\textbf{Oasst}} 
& \multicolumn{2}{c}{\textbf{Koala}}
& \multicolumn{2}{c}{\textbf{Self-instruct}}
& \multicolumn{2}{c}{\textbf{Average}}\\
\cmidrule(r){2-3}
\cmidrule(r){4-5}%
\cmidrule(r){6-7}%
\cmidrule(r){8-9}%
\cmidrule(r){10-11}%
\cmidrule(r){12-13}%
\multicolumn{1}{c}{~} 
& $\rho_{\text{non-trans}}^\downarrow$ & $\tau_{\text{avg}}^\downarrow$ 
& $\rho_{\text{non-trans}}^\downarrow$ & $\tau_{\text{avg}}^\downarrow$
& $\rho_{\text{non-trans}}^\downarrow$ & $\tau_{\text{avg}}^\downarrow$
& $\rho_{\text{non-trans}}^\downarrow$ & $\tau_{\text{avg}}^\downarrow$
& $\rho_{\text{non-trans}}^\downarrow$ & $\tau_{\text{avg}}^\downarrow$
& $\rho_{\text{non-trans}}^\downarrow$ & $\tau_{\text{avg}}^\downarrow$
\\
\midrule
\textbf{Stronger LLMs}\\
\midrule
Qwen2.5-Max   & 63.68\%  & 0.8047 &75.36\%  &0.8448 &64.29\%&0.7883&71.52\%&0.8296&64.97\%&0.7803 &67.96\%&0.8095\\ 
\midrule
\textbf{Base Model}\\
\midrule
LLaMA-Base  &76.41\%&0.8518&66.96\%&0.8461&67.40\%&0.7933&69.87\%&0.8180&71.03\%&0.8091 &70.33\% &0.8237\\ 
\midrule
\textbf{Cleaned group}\\
\midrule
LLaMA-Helpful\_Base-Cleaned &40.20\%  &0.6523 &47.86\% &0.6810 &49.92\% &0.6827 &50.92\%  &0.7003 &55.33\% &0.7300 &48.85\% &0.6893\\
LLaMA-Vicuna-Cleaned &49.83\%  &0.7139 &45.36\% &0.6910 &48.02\% &0.6863 &53.02\%  &0.7295 &57.14\% &0.7408 &50.67\% &0.7123\\
LLaMA-Oasst-Cleaned &50.28\%  &0.6655 &47.68\% &0.6812&43.01\% &0.6288 &50.73\%  &0.7034 &54.25\% &0.7056 &49.19\% &0.6769\\
LLaMA-Koala-Cleaned &44.85\%  &0.6588 &46.61\% &0.6826&44.38\% &0.6591 &45.51\%  &0.6827 &50.11\% &0.7031 &46.29\% &0.6773\\
LLaMA-Self-instruct-Cleaned &40.42\%  &0.6373 &49.29\% &0.6951 &41.95\% &0.6423 &48.99\%  &0.6959 &46.15\% &0.6652 &45.36\%&0.6672\\
\midrule
\textbf{Raw group}\\
\midrule
LLaMA-Helpful\_Base-Raw &59.03\%  &0.7654 &60.71\% &0.7888 &59.42\% &0.7682 &61.36\%  &0.7965 &63.78\% &0.7886&60.86\% &0.7815\\
LLaMA-Vicuna-Raw &56.92\%  &0.8028 &60.18\% &0.7718 &58.81\% &0.7646 &60.35\%  &0.7943 &59.69\% &0.7644 &59.19\%&0.7797 \\
LLaMA-Oasst-Raw &59.80\%  &0.7821 & 64.64\% &0.7878 &58.21\% &0.7596 &67.49\%  &0.8062 &63.89\% &0.7745 &62.81 &0.7820\\
LLaMA-Koala-Raw &58.03\%  &0.7532 & 53.75\% &0.7601 &57.22\% &0.7457 &57.23\%  &0.7597 &65.59\% &0.7887 &58.36\% &0.7615\\
LLaMA-Self-instruct-Raw &62.79\%  &0.7860 &64.11\% &0.7885 &59.95\% &0.7830 &65.02\%  &0.8054 &56.52\% &0.7559 &61.68\% &0.7838\\
\bottomrule
\end{tabular}
}
\caption{\label{tab:cross-LLaMA-result} 
Comparison of \textbf{Preference Non-Transitivity} and \textbf{Overall Clarity} for evaluator LLMs. 
LLaMA-Base refers to the original LLaMA3.1-8B-Instruct model. \textbf{Raw group} and \textbf{Cleaned group} refer to models fine-tuned on the original and filtered training sets, respectively. 
Specifically, LLaMA-Helpful\_Base-Raw is fine-tuned on the \textbf{``Raw"} training data generated from the \textbf{Helpful\_Base} dataset, while LLaMA-Helpful\_Base-Cleaned is fine-tuned on the corresponding \textbf{``Cleaned"} dataset after removing non-transitive preference data.
}
\end{table*}

\begin{table*}[htbp]
    \centering
    \begin{tabular}{|p{13cm}|}
        \hline
        \textbf{System}
        \\ \hline
You are a highly efficient assistant, who evaluates and selects the best large language model (LLMs) based on the quality of their responses to a given instruction. This process will be used to create a leaderboard reflecting the most accurate and human-preferred answers.\\ \hline
        \textbf{User}
        \\ \hline
I require a leaderboard for various large language models. I'll provide you with prompts given to these models and their corresponding outputs. Your task is to assess these responses and select the model that produces the best output from a human perspective.\newline

\#\# Instruction\newline

\{\newline
"instruction": """\{instruction\}""",\newline
\}

\#\# Model Outputs\newline

Here are the unordered outputs from the models. Each output is associated with a specific model, identified by a unique model identifier.\newline

\{\newline
    \{\newline
        "model\_identifier": "m",\newline
        "output": """\{output\_1\}"""\newline
    \},\newline
    \{\newline
        "model\_identifier": "M",\newline
        "output": """\{output\_2\}"""\newline
    \}\newline
\}\newline

\#\# Task\newline

Evaluate the models based on the quality and relevance of their outputs, and select the model that generated the best output. Answer by first providing a concise explanation and then end your answer by providing the model identifier of the best output. We will use the last character of your output `output[-1]` as the name of the best model, so make sure you finish with the token of the model identifiers and nothing else: `m` or `M` (no quotes, no dots, no backticks, no new lines, ...). For example:\newline

\#\#\# Concise explanation
...some text...\newline

\#\#\# Which is best, m or M?\newline
M

Now is your turn.\newline

\#\# Your answer: "Concise explanation" followed by "Which is best, m or M?"
\\ \hline
    \end{tabular}
    \caption{The Chain-of-Thought Comparison prompt for pairwise comparison.}
    \label{tab:prompts-cot}
\end{table*}

\begin{table*}[htbp]
    \centering
    \begin{tabular}{|p{13cm}|}
        \hline
        \textbf{System}
        \\ \hline
You are a highly efficient assistant, who evaluates and selects the best large language model (LLMs) based on the quality of their responses to a given instruction. This process will be used to create a leaderboard reflecting the most accurate and human-preferred answers.\\ \hline
        \textbf{User}
        \\ \hline
I require a leaderboard for various large language models. I'll provide you with prompts given to these models and their corresponding outputs. Your task is to assess these responses, and select the model that produces the best output from a human perspective. If you determine that both outputs are of equal quality or are unable to decide which one is better, you should indicate a tie by providing the identifier `D`.\newline

\#\# Instruction\newline

\{\newline
"instruction": """\{instruction\}""",\newline
\}

\#\# Model Outputs\newline

Here are the unordered outputs from the models. Each output is associated with a specific model, identified by a unique model identifier.\newline

\{\newline
    \{\newline
        "model\_identifier": "m",\newline
        "output": """\{output\_1\}"""\newline
    \},\newline
    \{\newline
        "model\_identifier": "M",\newline
        "output": """\{output\_2\}"""\newline
    \}\newline
\}\newline

\#\# Task\newline

Evaluate the models based on the quality and relevance of their outputs, and select the model that generated the best output. Answer by first providing a concise explanation and then end your answer by providing the model identifier of the best output. If you determine that both outputs are of equal quality or cannot decide which one is better, indicate a tie by using the identifier `D`. We will use the last character of your output `output[-1]` as the name of the best model, so make sure you finish with the token of the model identifiers and nothing else: `m`, `M`  or `D` (no quotes, no dots, no backticks, no new lines, ...). For example:\newline

\#\#\# Concise explanation
...some text...\newline

\#\#\# Which is best, m, M or D?\newline
M

Now is your turn.\newline

\#\# Your answer: "Concise explanation" followed by "Which is best, m, M or D?"
\\ \hline
    \end{tabular}
    \caption{The Chain-of-Thought Comparison prompt (Tie Allowed) for pairwise comparison.}
    \label{tab:prompts-cot-allow-tie}
\end{table*}

\end{document}